\documentclass[runningheads]{llncs}
\usepackage[T1]{fontenc}
\usepackage{amsmath,amssymb,bm}
\usepackage{graphicx}
\usepackage{booktabs}
\usepackage{hyperref}
\usepackage{xcolor}
\usepackage{multirow}

\begin{document}

\title{Bag-of-Visual-Words for Spatial Mapping of Lung Adenocarcinoma Growth Patterns\thanks{Supported by the Swiss National Science Foundation (SNSF)
under Grant No.~220761}}

\titlerunning{BoVW Spatial Mapping of LUAD Patterns}

\author{Darya Ardan\inst{1,2} \and
Valentin Oreiller\inst{2} \and
Henning Müller\inst{1,3}}
\authorrunning{D. Ardan, V. Oreiller et al.}
\institute{Department of Computer Science, University of Geneva, Geneva, Switzerland \and
Informatics Institute, University of Applied Sciences Western Switzerland (HES-SO Valais), Sierre, Switzerland \and
Radiology Service, Medical Faculty, University of Geneva, Geneva, Switzerland
}
\maketitle

\begin{abstract}
Spatial mapping of lung adenocarcinoma (LUAD) growth patterns
across whole slide images (WSIs) requires resolving
architectural context at the region level, yet existing
methods operate at the individual tile level and produce
generic morphological clusters rather than clinically defined
pattern maps.
We propose a weakly supervised Bag-of-Visual-Words (BoVW) pipeline that learns a visual vocabulary from frozen foundation model embeddings extracted from a small set of annotated regions of interest (ROIs). Pattern prototypes are constructed as mean BoVW histograms of same-label ROIs and used for nearest-prototype classification of sliding-window regions under Jensen--Shannon divergence. The resulting predictions are projected onto the WSI tile grid to produce interpretable spatial pattern maps.
We evaluate the method on 87 CPTAC-LUAD patients using three foundation model encoders and multiple vocabulary sizes on two clinically motivated tasks. For tumour/healthy classification, the best configuration achieves a balanced accuracy of $0.974$ with H-Optimus-1, approaching the $0.987$ obtained by a supervised SVM trained on mean-pooled WSI embeddings. For binary histologic grade classification, the BoVW pipeline achieves higher balanced accuracy than the supervised baseline for all encoders, suggesting that ROI-level pattern decomposition preserves grade-relevant heterogeneity that is attenuated by global mean pooling.
\end{abstract}

\section{Introduction}\label{sec:intro}

Lung adenocarcinoma (LUAD) is characterised by a diverse set of histological growth patterns, namely lepidic, acinar, papillary, micropapillary, and solid, whose spatial distribution within a tumour carries significant prognostic information~\cite{moreira2020}. In clinical practice, pathologists report the dominant growth pattern per whole slide image (WSI), a convention that obscures the intratumoural heterogeneity that is increasingly recognised as a determinant of disease progression and treatment response. Spatial maps showing the distribution of growth patterns across a WSI would provide clinicians with richer information than a single dominant label, enabling quantitative characterisation of pattern composition and heterogeneity at the slide level. Yet the computational tools to generate such maps remain limited.

The challenge concerns both supervision and spatial scale. Prior supervised approaches have used pixel-wise semantic segmentation~\cite{anorak} or patch-level classification across WSIs~\cite{wei2019,lami2023}. However, dense pixel annotations are costly and arguably finer than required for regionally defined LUAD growth patterns, while independent patch predictions capture only limited architectural context. Existing datasets such as ANORAK~\cite{anorak} instead annotate selected ROIs, whereas larger weakly labelled cohorts typically provide only a dominant slide label. These limitations motivate ROI-level spatial mapping from limited regional supervision.


Unsupervised methods have emerged as a promising alternative. Histomorphological Phenotype Learning (HPL)~\cite{quiros2024hpl} and PANTHER~\cite{song2024panther} demonstrate that self-supervised tile embeddings can be clustered into morphologically coherent phenotype groups, producing spatial maps of tissue composition without any annotation. However, these methods operate at the individual tile level, typically $224{\times}224$ pixels, and discover generic morphological concepts rather than the specific WHO-defined LUAD growth patterns. This is a fundamental limitation: LUAD growth patterns are architectural phenomena defined by the spatial arrangement of cells across multiple tile widths. A lepidic pattern is recognisable from the preservation of alveolar architecture across a region, not from any single tile in isolation. Tile-level clustering conflates locally similar tiles that belong to architecturally distinct patterns, reducing discriminability for the clinically relevant pattern taxonomy.

We adopt the region of interest (ROI) as the unit of
analysis for LUAD growth pattern mapping.
This is motivated by how pathologists annotate these
patterns in practice: annotation protocols instruct
experts to delineate regions as large as possible that
are relatively pure, containing a single
pattern~\cite{reisenbuchler2025cellomaps2,anorak}.
Furthermore, growth patterns are defined by the spatial
arrangement of cells across a tissue region rather than
by local pixel features~\cite{reisenbuchler2025cellomaps2},
and empirical studies show that pattern recognition
benefits from a field of view spanning several hundred
microns~\cite{reisenbuchler2025cellomaps2}.
An ROI spanning multiple tile widths therefore captures
the architectural context required to identify each
pattern, while remaining far coarser than pixel-level
segmentation and computationally tractable at the WSI
scale.

In this paper we propose a weakly supervised pipeline for
spatial LUAD growth pattern mapping that operates at the
ROI level. A visual vocabulary is learned by clustering tile embeddings
from 168 annotated ROIs using a frozen foundation model encoder, and per-pattern prototypes are constructed as mean BoVW histograms of same-label ROIs.
Pattern assignment on unseen WSIs is performed by
nearest-prototype retrieval under Jensen--Shannon Divergence,
and predictions are assembled into spatial pattern maps on
the WSI tile grid.
Evaluated on CPTAC-LUAD WSIs on tumour/healthy
classification and binary histologic grade classification,
our method achieves balanced accuracies of $0.974$ and
$0.729$ respectively, with the latter outperforming a
directly supervised baseline ($0.678$).

\section{Method}\label{sec:method}

\begin{figure}[t]
  \centering
  \includegraphics[width=\textwidth]{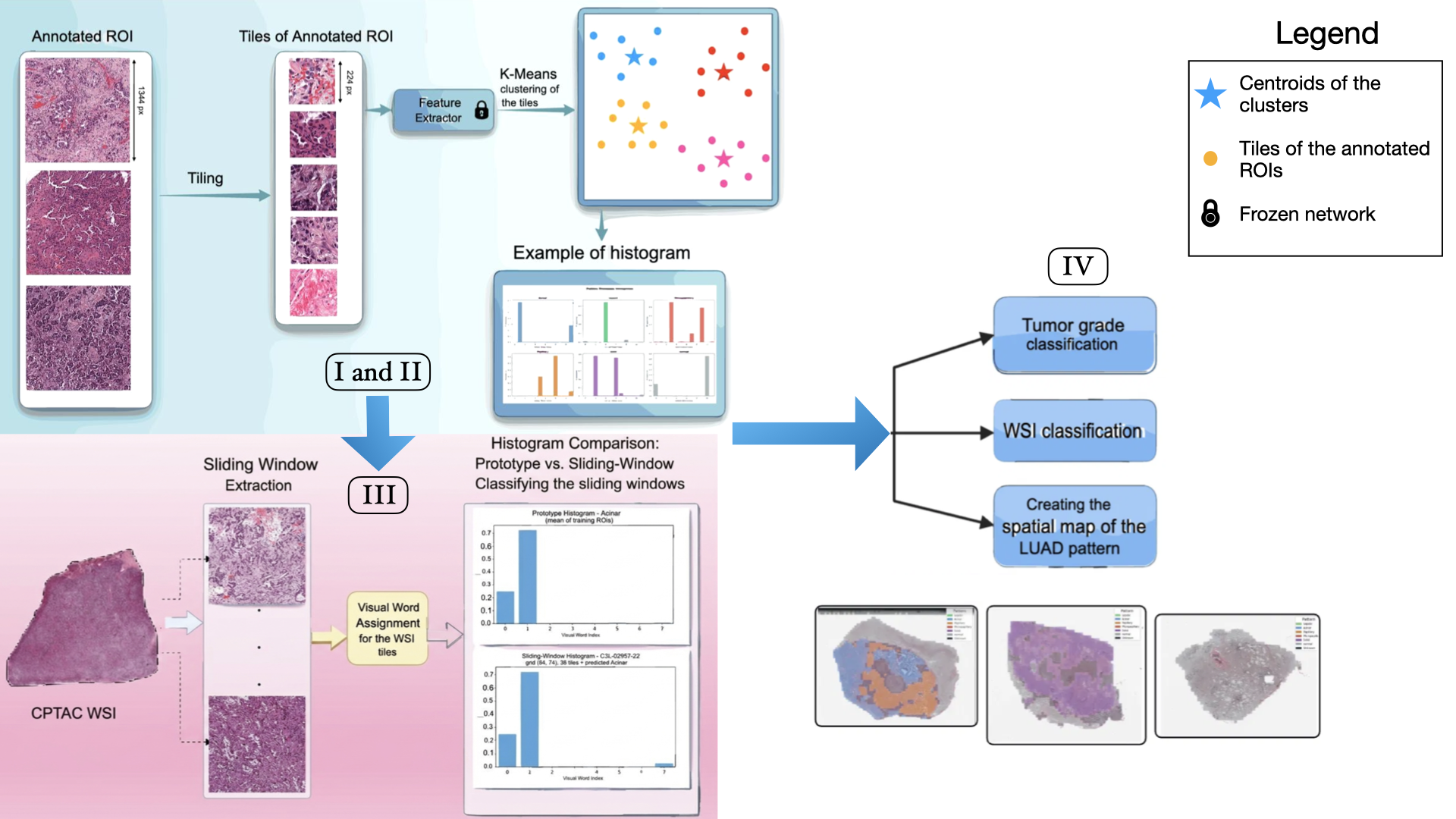}
  \caption{Overview of the proposed pipeline.
         \textbf{(I, II)}: Visual vocabulary learning from
         annotated ROIs via a frozen foundation
         model encoder, and BoVW histogram encoding.
         \textbf{(III)}: Sliding-window BoVW encoding of
         WSIs and nearest-prototype classification
         under JSD.
         \textbf{(IV)}: Downstream outputs: tumour/healthy
         classification, histologic grade classification,
         and spatial LUAD pattern maps.}
  \label{fig:pipeline}
\end{figure}

We propose a weakly supervised pipeline for spatially mapping
LUAD growth patterns across WSIs.
Given a small set of pattern-labelled regions of interest (ROIs)
and a collection of unannotated whole slide images (WSIs),
the goal is to assign a growth pattern label to every spatial
position of each WSI without tile-level or pixel-level
supervision.
The pipeline proceeds in four stages:
(\textit{i})~visual vocabulary learning from annotated ROI tiles,
(\textit{ii})~ROI histogram encoding,
(\textit{iii})~per-pattern prototype construction, and
(\textit{iv})~WSI pattern assignment and spatial mapping,
as illustrated in Figure~\ref{fig:pipeline}.

\subsection{Visual Vocabulary Learning}\label{sec:vocab}

Let $\{\mathbf{x}_i\}_{i=1}^{N} \subset \mathbb{R}^{d}$ denote
the set of $\ell_2$-normalised tile embeddings extracted from
all annotated ROIs using a frozen foundation model
encoder $f_\theta$.
Although individual tiles carry no label, they are drawn
exclusively from tissue regions whose enclosing ROI has been
assigned a pattern label.

A visual vocabulary of $k$ visual words is learned by applying
$K$-Means to the embeddings of tiles extracted from the
annotated ROIs (stages~I and II in Figure~\ref{fig:pipeline}):
\begin{equation}
  \min_{\mathcal{C}}
  \sum_{i=1}^{N}
  \min_{\mathbf{c} \in \mathcal{C}}
  \|\mathbf{x}_i - \mathbf{c}\|_2^2,
  \quad
  \mathcal{C} = \{\mathbf{c}_1, \ldots, \mathbf{c}_k\},
  \label{eq:kmeans}
\end{equation}
yielding $k$ centroids $\mathcal{C}$ that constitute the
vocabulary.
The algorithm operates without access to pattern labels,
making vocabulary construction entirely unsupervised.

\subsection{ROI Histogram Encoding}\label{sec:histogram}

Given the vocabulary $\mathcal{C}$ of $k$ centroids, each
annotated ROI is encoded as a normalised histogram over visual
words, the Bag-of-Visual-Words (BoVW)
representation~\cite{sivic2003,caicedo2009}.
For the $n$ tile embeddings
$\{\mathbf{x}_j\}_{j=1}^{n}$ belonging to ROI $r$, each tile
is assigned to its nearest centroid:
\begin{equation}
  z_j =
  \operatorname*{arg\,min}_{z \in \{1,\ldots,k\}}
  \|\mathbf{x}_j - \mathbf{c}_z\|_2^2.
  \label{eq:assign}
\end{equation}
Because all embeddings and centroids are $\ell_2$-normalised
and therefore lie on the unit hypersphere, minimising squared
Euclidean distance is equivalent to maximising cosine
similarity: $\|\mathbf{x} - \mathbf{c}\|_2^2 = 2(1 -
\mathbf{x}^\top \mathbf{c})$.
The assignment in Eq.~\eqref{eq:assign} thus selects the
visually most similar visual word for each tile.

The ROI histogram is formed as the mean of one-hot
assignment vectors:
\begin{equation}
  \mathbf{h}_r
  = \frac{1}{n} \sum_{j=1}^{n} \mathbf{e}_{z_j}
  \;\in\; \mathbb{R}^{k},
  \label{eq:histogram}
\end{equation}
where $\mathbf{e}_{z_j} \in \{0,1\}^k$ is the one-hot
indicator of word $z_j$.
By construction $\mathbf{h}_r$ is $\ell_1$-normalised and
constitutes a valid discrete probability distribution over
the $k$ visual words.

\subsection{Per-Pattern Prototype Construction}\label{sec:proto}

Let $\mathcal{R}_g$ denote the set of BoVW histograms of all
annotated ROIs carrying label $g$, where
$g \in \mathcal{G} = \{$Lepidic, Acinar, Papillary,
Micropapillary, Solid, Normal$\}$.
The prototype of pattern $g$ is the mean BoVW histogram
over all ROIs of that pattern:
\begin{equation}
  \bm{\mu}_g
  = \frac{1}{|\mathcal{R}_g|}
    \sum_{\mathbf{h} \in \mathcal{R}_g} \mathbf{h},
  \label{eq:prototype}
\end{equation}
Because each histogram is already $\ell_1$-normalised,
$\bm{\mu}_g$ is itself a valid probability distribution.

\paragraph{Vocabulary quality criterion.}
The separability of the learned prototypes is assessed by
computing the full $|\mathcal{G}| \times |\mathcal{G}|$
matrix of pairwise Jensen--Shannon Divergences (JSD)
between all prototype pairs.
For two distributions $P, Q \in \mathbb{R}^k$:
\begin{equation}
  \operatorname{JSD}(P \| Q)
  = \frac{1}{2} D_{\mathrm{KL}}\!\left(P \,\Big\|\, M\right)
  + \frac{1}{2} D_{\mathrm{KL}}\!\left(Q \,\Big\|\, M\right),
  \quad
  M = \tfrac{1}{2}(P + Q),
  \label{eq:jsd}
\end{equation}
where
$D_{\mathrm{KL}}(P \| Q)
 = \sum_{l=1}^{k} P[l] \log \frac{P[l]}{Q[l]}$
is the KL divergence.
JSD is symmetric, bounded in $[0,1]$, and well-defined even
when histograms contain zero-count bins (a small smoothing
constant $\varepsilon$ is applied).
A vocabulary passes quality control if
$\min_{g \neq g'} \operatorname{JSD}(\bm{\mu}_g, \bm{\mu}_{g'})
 > \delta$, where $\delta = 0.05$ is the threshold below which
two prototypes are considered effectively indistinguishable.

\subsection{WSI Pattern Assignment and Spatial Mapping}
\label{sec:mapping}

\paragraph{Sliding-window ROI extraction.}
Each WSI is tessellated into overlapping sliding-window ROIs
(stage~III in Figure~\ref{fig:pipeline}).
A window of $W \times W$ tiles is slid with stride $s$ tiles
across the tile grid; windows with fewer than $\theta_{\min}$
valid (non-background) tiles are discarded.
Each surviving window is encoded into a BoVW histogram
$\mathbf{h}_{r}$ using
Eqs.~\eqref{eq:assign}--\eqref{eq:histogram}.

\paragraph{Pattern assignment.}
Each WSI-ROI is assigned the pattern whose prototype is
closest under JSD (Eq.~\eqref{eq:jsd}):
\begin{equation}
  \hat{g}_r =
  \operatorname*{arg\,min}_{g \in \mathcal{G}}\;
  \operatorname{JSD}\!\left(\mathbf{h}_{r} \,\|\, \bm{\mu}_g\right).
  \label{eq:classify}
\end{equation}

\paragraph{Spatial mapping.}
ROI predictions are projected back onto the WSI tile grid
(stage~IV in Figure~\ref{fig:pipeline}).
Classification confidence for each ROI is defined as the
normalised margin between the JSD distances to the closest
and second-closest prototypes, $\kappa_r = 1 - d_1 / d_2$,
where $d_1$ and $d_2$ denote the closest and second-closest
prototype distances respectively.
ROIs whose confidence falls below threshold $\tau$ are
labelled \textit{Unknown}; each \textit{Unknown} ROI is
then relabelled by majority vote over its immediate grid
neighbours, exploiting the spatial contiguity of LUAD
growth patterns.

\paragraph{WSI-level classification tasks.}
Two downstream tasks are evaluated.

\textit{Task~1: tumour/healthy classification.}
A WSI is classified as tumour if at least one ROI prediction
belongs to a cancer pattern
($\mathcal{G} \setminus \{\text{Normal}\}$);
otherwise it is classified as healthy.

\textit{Task~2: binary histologic grade classification.}
High-grade patterns (Solid, Micropapillary) are associated
with worse survival and higher recurrence risk than
low-grade patterns (Lepidic, Acinar,
Papillary)~\cite{moreira2020,travis2015,sica2010}.
Each patient is assigned a binary grade from their
ROI-level predictions using a highest-grade-wins rule,
consistent with IASLC recommendations~\cite{moreira2020}:
\begin{equation}
  \hat{g}_{\text{patient}}
  = \begin{cases}
      \text{high-grade} &
        \text{if } \exists\, r : \hat{g}_r \in
        \{\text{Solid, Micropapillary}\}, \\
      \text{low-grade}  & \text{otherwise.}
    \end{cases}
  \label{eq:grade}
\end{equation}

\section{Experiments and Results}\label{sec:results}
We evaluate the pipeline on 87 CPTAC-LUAD~\cite{cptac} patients comprising 453 WSI entries, using H-Optimus-1~\cite{hoptimus1}, H0-Mini~\cite{hoptimus0mini}, and UNI2-h~\cite{chen2024uni}, with vocabulary sizes $k \in {6,8,12,16,18}$. All analyses use $224{\times}224$-pixel tiles at $20{\times}$ magnification, matching the native input configuration of the encoders.

The reference pool contains 168 pattern-enriched ROIs of $1344{\times}1344$ pixels, corresponding to 6,048 tiles. ROIs were selected from CPTAC-LUAD slides with dominant-pattern metadata. A pretrained tumour-localisation model was used only to identify tumour tissue, after which ROIs were manually selected by a computational pathology researcher without formal pathology training using the slide metadata and published morphological criteria. These ROIs are used exclusively for vocabulary learning and prototype construction. Evaluation patients are disjoint from those contributing reference ROIs.

Sliding-window regions extracted during WSI inference use the same  size $1344{\times}1344$ pixel field of view. Results are reported for WSI-level tumour/healthy classification and patient-level binary histologic grade classification. 
The supervised baseline mean-pools tile embeddings within each region and region embeddings within each WSI, followed by an SVM evaluated using leave-one-patient-out cross-validation. In contrast, the BoVW prototypes are learned from the disjoint reference ROI pool and applied without fitting to labels from the evaluation cohort.
Attention-based multiple instance learning~\cite{ilse2018abmil} was not evaluated because the labelled cohort was considered insufficient for reliable training.

\subsection{Vocabulary Evaluation}\label{sec:vocab_eval}

\paragraph{Inter-prototype separability.}
Table~\ref{tab:jsd_h1} reports the inter-prototype JSD matrix
for H-Optimus-1 at $k{=}8$.
The lowest separation is between Acinar and Papillary, reflecting their known
architectural similarity as glandular patterns~\cite{thunnissen2012}.
The highest separation is between Micropapillary and Lepidic, consistent with these patterns representing opposite ends of the LUAD architectural spectrum.
Table~\ref{tab:jsd_summary} shows that H-Optimus-1 achieves
consistently higher minimum JSD than H0-Mini, while UNI2-h
attains the highest absolute minimum yet underperforms on
downstream classification (Tables~\ref{tab:task1_full}
and~\ref{tab:task2_full}), suggesting prototype separability
is necessary but not sufficient for WSI-level pattern mapping.
At $k{=}6$, vocabulary size equals the number of pattern
classes, yielding $\mathrm{JSD} = 1.000$ by construction
due to trivially disjoint prototypes; these values are
excluded from the trend analysis.
For all encoders, minimum JSD increases monotonically with
$k$, yet the vocabulary size that maximises separability
differs from the one that maximises classification accuracy.
Every (encoder, $k$) configuration evaluated satisfies the
quality criterion of Section~\ref{sec:proto}
($\min_{g \neq g'} \mathrm{JSD}(\bm{\mu}_g, \bm{\mu}_{g'}) > \delta = 0.05$,
excluding the degenerate $k{=}6$ case), so no vocabulary was
excluded from the downstream evaluation on this basis.

\begin{table}[t]
\centering
\caption{Inter-prototype JSD matrix for H-Optimus-1, $k{=}8$.
         All off-diagonal values exceed $0.74$, confirming
         strong prototype separability across all pattern pairs.}
\label{tab:jsd_h1}
\setlength{\tabcolsep}{4pt}
\small
\begin{tabular}{lcccccc}
\toprule
& \textbf{Acin.} & \textbf{Lepi.} & \textbf{Microp.}
& \textbf{Pap.} & \textbf{Solid} & \textbf{Norm.} \\
\midrule
\textbf{Acinar}    & ---   & 0.994 & 0.940 & 0.741 & 0.746 & 0.976 \\
\textbf{Lepidic}   & 0.994 & ---   & 0.996 & 0.855 & 0.993 & 0.967 \\
\textbf{Micropap.} & 0.940 & 0.996 & ---   & 0.924 & 0.940 & 0.945 \\
\textbf{Papillary} & 0.741 & 0.855 & 0.924 & ---   & 0.955 & 0.991 \\
\textbf{Solid}     & 0.746 & 0.993 & 0.940 & 0.955 & ---   & 0.929 \\
\textbf{Normal}    & 0.976 & 0.967 & 0.945 & 0.991 & 0.929 & ---   \\
\bottomrule
\end{tabular}
\end{table}

\begin{table}
\centering
\caption{Minimum off-diagonal inter-prototype JSD.
         Higher values indicate more separable prototype pairs.
         }
\label{tab:jsd_summary}
\setlength{\tabcolsep}{5pt}
\small
\begin{tabular}{lccccc}
\toprule
\textbf{Encoder} & $k{=}6$ & $k{=}8$ & $k{=}12$
                 & $k{=}16$ & $k{=}18$ \\
\midrule
H-Optimus-1      & $1.000^\dagger$ & 0.741 & 0.752 & 0.785 & 0.825 \\
H0-Mini & $1.000^\dagger$ & 0.469          & 0.516 & 0.506 & 0.545 \\
UNI2-h             & $1.000^\dagger$ & 0.824          & 0.844 & 0.830 & 0.910 \\
\bottomrule
\multicolumn{6}{l}{$^\dagger$Degenerate case: disjoint support by construction.}
\end{tabular}
\end{table}

\subsection{Spatial Pattern Maps}\label{sec:spatial_maps}

\begin{figure}[t]
  \centering
  \includegraphics[width=\textwidth]{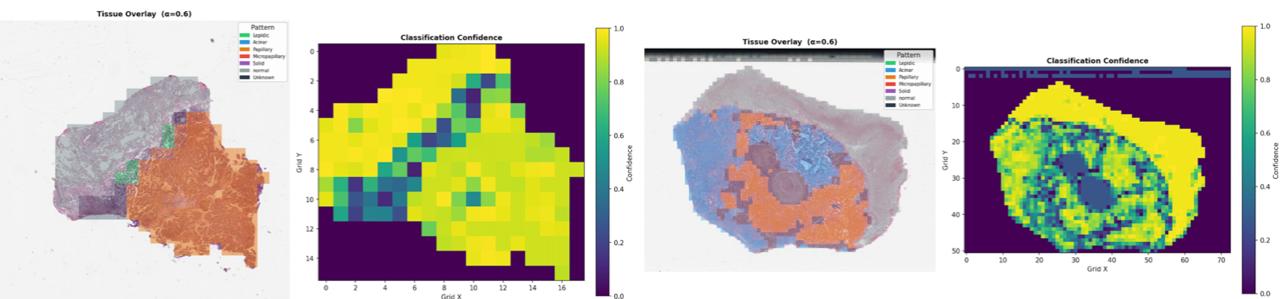}
  \caption{Spatial pattern maps for two CPTAC-LUAD WSIs
           (H-Optimus-1, $k{=}8$). Pattern overlay (left)
           shows per-ROI assignments superimposed on the
           tissue thumbnail. Confidence heatmap (right)
           encodes classification confidence, with yellow
           indicating high confidence and dark purple
           indicating low confidence or \textit{Unknown}
           regions.}
  \label{fig:spatial_maps}
\end{figure}

Each sliding-window ROI is coloured by $\hat{g}_r$ and placed
at its centre coordinates in tile units, with the confidence
threshold set to $\tau = 0.5$ and confidence
$\kappa_r$ encoded in a
complementary heatmap.
Representative maps for two CPTAC-LUAD slides are shown in
Figure~\ref{fig:spatial_maps} (H-Optimus-1, $k{=}8$).
Both slides show spatially coherent pattern regions consistent
with expected tissue architecture, and uncertainty
concentrates at pattern boundaries rather than being randomly
distributed. This qualitative visual inspection suggests
that the pipeline recovers plausible pattern organisation;
a formal assessment by pathologists is left to future work
(see Section~\ref{sec:discussion}).

\subsection{Downstream Classification}\label{sec:downstream}

\paragraph{Task~1: tumour/healthy classification.}
The BoVW pipeline using H-Optimus-1 and H0-Mini both peak at $k{=}16$, while UNI2-h improves monotonically with $k$, peaking at $k{=}18$, substantially below the other two encoders (Table~\ref{tab:task1_full}). The small gap between the supervised baseline and the best BoVW configurations (${\leq}0.013$ for H-Optimus-1)
indicates that the training-free prototype pipeline achieves
near-parity with a directly supervised method for tumour
detection, a task for which the embedding signal is strong
across all encoders.

\paragraph{Task~2: binary histologic grade classification.}
The BoVW pipeline achieves higher balanced accuracy than the supervised SVM baseline for all three encoders (Table~\ref{tab:task2_full}), with H-Optimus-1 at $k{=}8$ yielding the highest BoVW performance.
This consistent advantage across encoders demonstrates that
distributing WSI representation over pattern-relevant visual
words preserves grade-discriminative spatial heterogeneity
that is lost when embeddings are mean-pooled to a single
vector.
Notably, the optimal $k$ for Task~2 ($k{=}8$) is coarser
than for Task~1 ($k{=}16$), indicating that grade
discrimination and tumour detection impose different
requirements on vocabulary granularity. Task~2 results should be interpreted with caution given the
cohort imbalance (68 low-grade, 19 high-grade), and
validated on larger, balanced datasets.

\begin{table}
\centering
\caption{Task~1 (tumor vs healthy). $k$ is vocabulary size.
Baseline: SVM on mean-pooled ROI embeddings without vocabulary.}
\label{tab:task1_full}
\setlength{\tabcolsep}{3pt}
\scriptsize
\begin{tabular}{llc ccc ccc}
\toprule
& & & \multicolumn{3}{c}{\textbf{Healthy}}
  & \multicolumn{3}{c}{\textbf{Tumour}} \\
\cmidrule(lr){4-6}\cmidrule(lr){7-9}
\textbf{Encoder} & $k$ & \textbf{Bal.Acc}
  & P & R & F1 & P & R & F1 \\
\midrule
\multirow{6}{*}{H-Optimus-1}
  & 6           & 0.738          & 1.00 & 0.48 & 0.64 & 0.77 & 1.00 & 0.87 \\
  & 8           & 0.955          & 0.99 & 0.91 & 0.95 & 0.95 & 1.00 & 0.97 \\
  & 12          & 0.765          & 1.00 & 0.53 & 0.69 & 0.79 & 1.00 & 0.88 \\
  & \textbf{16} & \textbf{0.974} & 0.99 & 0.95 & 0.97 & 0.97 & 1.00 & 0.98 \\
  & 18          & 0.795          & 0.99 & 0.59 & 0.74 & 0.81 & 1.00 & 0.90 \\
  \cmidrule(lr){2-9}
  & Baseline    & 0.987          & 0.99 & 0.98 & 0.98 & 0.99 & 0.99 & 0.99 \\
\midrule
\multirow{6}{*}{H0-Mini}
  & 6           & 0.818          & 1.00 & 0.64 & 0.78 & 0.83 & 1.00 & 0.91 \\
  & 8           & 0.827          & 1.00 & 0.65 & 0.79 & 0.84 & 1.00 & 0.91 \\
  & 12          & 0.858          & 0.98 & 0.72 & 0.83 & 0.87 & 0.99 & 0.92 \\
  & \textbf{16} & \textbf{0.955} & 0.99 & 0.91 & 0.95 & 0.95 & 1.00 & 0.97 \\
  & 18          & 0.921          & 0.99 & 0.85 & 0.91 & 0.92 & 1.00 & 0.96 \\
  \cmidrule(lr){2-9}
  & Baseline    & 0.981          & 0.98 & 0.98 & 0.98 & 0.99 & 0.99 & 0.99 \\
\midrule
\multirow{6}{*}{UNI2-h}
  & 6           & 0.698          & 1.00 & 0.40 & 0.57 & 0.75 & 1.00 & 0.86 \\
  & 8           & 0.694          & 1.00 & 0.39 & 0.56 & 0.75 & 1.00 & 0.85 \\
  & 12          & 0.775          & 1.00 & 0.55 & 0.71 & 0.80 & 1.00 & 0.89 \\
  & 16          & 0.788          & 0.99 & 0.58 & 0.73 & 0.81 & 1.00 & 0.89 \\
  & \textbf{18} & \textbf{0.810} & 0.99 & 0.62 & 0.77 & 0.83 & 1.00 & 0.90 \\
  \cmidrule(lr){2-9}
  & Baseline    & 0.984          & 0.99 & 0.98 & 0.98 & 0.99 & 0.99 & 0.99 \\
\bottomrule
\end{tabular}
\end{table}


\begin{table}
\centering
\caption{Task~2 (low vs high grade). $k$ is vocabulary size.
Baseline: SVM on mean-pooled ROI embeddings without vocabulary.
}
\label{tab:task2_full}
\setlength{\tabcolsep}{3pt}
\scriptsize
\begin{tabular}{llcc ccc ccc}
\toprule
& & & & \multicolumn{3}{c}{\textbf{Low-grade}}
  & \multicolumn{3}{c}{\textbf{High-grade}} \\
\cmidrule(lr){5-7}\cmidrule(lr){8-10}
\textbf{Encoder} & $k$ & \textbf{Bal.Acc} & \textbf{Mac.F1}
  & P & R & F1 & P & R & F1 \\
\midrule
\multirow{6}{*}{H-Optimus-1}
  & 6           & 0.707          & 0.641          & 0.90 & 0.68 & 0.77 & 0.39 & 0.74 & 0.51 \\
  & \textbf{8}  & \textbf{0.729} & \textbf{0.671} & 0.91 & 0.72 & 0.80 & 0.42 & 0.74 & \textbf{0.54} \\
  & 12          & 0.714          & 0.651          & 0.90 & 0.69 & 0.78 & 0.40 & 0.74 & 0.52 \\
  & 16          & 0.714          & 0.651          & 0.90 & 0.69 & 0.78 & 0.40 & 0.74 & 0.52 \\
  & 18          & 0.714          & 0.651          & 0.90 & 0.69 & 0.78 & 0.40 & 0.74 & 0.52 \\
  \cmidrule(lr){2-10}
  & Baseline    & 0.678          & 0.685          & 0.86 & 0.88 & 0.87 & 0.53 & 0.47 & 0.50 \\
\midrule
\multirow{6}{*}{H0-Mini}
  & 6           & 0.685          & 0.612          & 0.90 & 0.63 & 0.74 & 0.36 & 0.74 & 0.48 \\
  & \textbf{8}  & \textbf{0.692} & \textbf{0.622}          & 0.90 & 0.65 & 0.75 & 0.37 & 0.74 & 0.49 \\
  & 12          & 0.673          & 0.615          & 0.88 & 0.66 & 0.76 & 0.36 & 0.68 & 0.47 \\
  & 16          & 0.651          & 0.586          & 0.88 & 0.62 & 0.72 & 0.33 & 0.68 & 0.45 \\
  & 18          & 0.666          & 0.605          & 0.88 & 0.65 & 0.75 & 0.35 & 0.68 & 0.46 \\
  \cmidrule(lr){2-10}
  & Baseline    & 0.644          & 0.650          & 0.84 & 0.87 & 0.86 & 0.47 & 0.42 & 0.44 \\
\midrule
\multirow{6}{*}{UNI2-h}
  & 6           & 0.699          & 0.631          & 0.90 & 0.66 & 0.76 & 0.38 & 0.74 & 0.50 \\
  & 8           & 0.692          & 0.622          & 0.90 & 0.65 & 0.75 & 0.37 & 0.74 & 0.49 \\
  & 12          & 0.699          & 0.631          & 0.90 & 0.66 & 0.76 & 0.38 & 0.74 & 0.50 \\
  & \textbf{16} & \textbf{0.707} & \textbf{0.641} & 0.90 & 0.68 & 0.77 & 0.39 & 0.74 & 0.51 \\
  & 18          & 0.707          & 0.641          & 0.90 & 0.68 & 0.77 & 0.39 & 0.74 & 0.51 \\
  \cmidrule(lr){2-10}
  & Baseline    & 0.693          & 0.708          & 0.86 & 0.91 & 0.89 & 0.60 & 0.47 & 0.53 \\
\bottomrule
\end{tabular}
\end{table}

\section{Discussion}\label{sec:discussion}

Our results show that a training-free BoVW pipeline anchored to a small set of reference ROIs can produce spatially coherent LUAD growth-pattern maps and competitive downstream classification. For binary grade classification, the BoVW pipeline achieves higher balanced accuracy than the supervised SVM baseline across all three encoders. This comparison should nevertheless be interpreted with care, since the SVM is trained through leave-one-patient-out cross-validation on the evaluation cohort, whereas the BoVW prototypes are learned from a disjoint reference set. The results suggest that decomposing each WSI into pattern-relevant visual words may preserve the contribution of spatially limited high-grade regions that can be attenuated by global mean pooling.


The two tasks benefit from different levels of vocabulary
granularity, reflecting the different nature of the
discriminative signal: separating tumour from healthy
tissue relies on coarser tissue-type differences, while
grade discrimination requires the vocabulary to resolve
finer architectural distinctions between pattern groups.

\paragraph{Limitations.}
Two data constraints limit the current study.
The annotated ROI pool is small and class-imbalanced
(14 to 44 ROIs per pattern), which may limit prototype
quality for rare patterns.
The evaluation dataset carries weak patient-level grade
labels with no WSI-level ground truth and pronounced class
imbalance (68 low-grade, 19 high-grade), which motivated
the choice of binary rather than per-class evaluation and
limits the conclusions that can be drawn from high-grade
F1 values.
Despite these constraints, the consistent advantage of the
BoVW pipeline over the supervised baseline across all
encoders and the qualitative coherence of the spatial maps
suggest that the method is promising.
Future work will leverage higher-quality hospital data
to enable a more rigorous evaluation of per-class spatial
mapping accuracy.


 
\bibliographystyle{splncs04}

\end{document}